\documentclass[letter, 10 pt, conference
]{ieeeconf}

\IEEEoverridecommandlockouts                              

\usepackage{amsmath}
\usepackage{amssymb}
\usepackage{mathtools}
\usepackage{graphicx}
\usepackage{float}
\usepackage{array}
\usepackage{tikz}
\usepackage{listings}
\usepackage{soul}
\usepackage{multirow}
\usepackage{array}

\PassOptionsToPackage{hyphens}{url}
\usepackage{hyperref}

\usepackage{float}

\usepackage{graphicx}
\usepackage{cite}
\usepackage{dsfont}
\usepackage{mathtools,etoolbox}
\usepackage[ruled, linesnumbered]{algorithm2e}
\usepackage[none]{hyphenat}
\usepackage[utf8]{inputenc}
\usepackage[english]{babel}

\usepackage{amsthm}
\usepackage{amsmath}
\usepackage{xfrac}
\usepackage{nicefrac}
\usepackage{booktabs}
\usepackage[switch, pagewise]{lineno}
\usepackage{nicefrac}
\usepackage{flushend}
\usepackage{xcolor}

\usepackage{multirow}
\usepackage{adjustbox} 
\usepackage{caption,setspace}

\hypersetup{colorlinks=true,urlcolor=red,citecolor=red}

\usepackage[placement=top]{background}

\backgroundsetup{
        placement=top,
        position={8cm,2cm}}
\usepackage[final]{pdfpages}

\title{\LARGE \bf Active Human Pose Estimation {\color{black}via an Autonomous UAV Agent}}

\author{Jingxi Chen, Botao He, Chahat Deep Singh, Cornelia Ferm{\"u}ller, Yiannis Aloimonos \\
Perception and Robotics Group, University of Maryland - College Park
}

\begin{document}
\makeatletter
\g@addto@macro\@maketitle{
\begin{figure}[H]
  \setlength{\linewidth}{\textwidth}
  \setlength{\hsize}{\textwidth}
    \centering
    \includegraphics[width=1.0\textwidth]{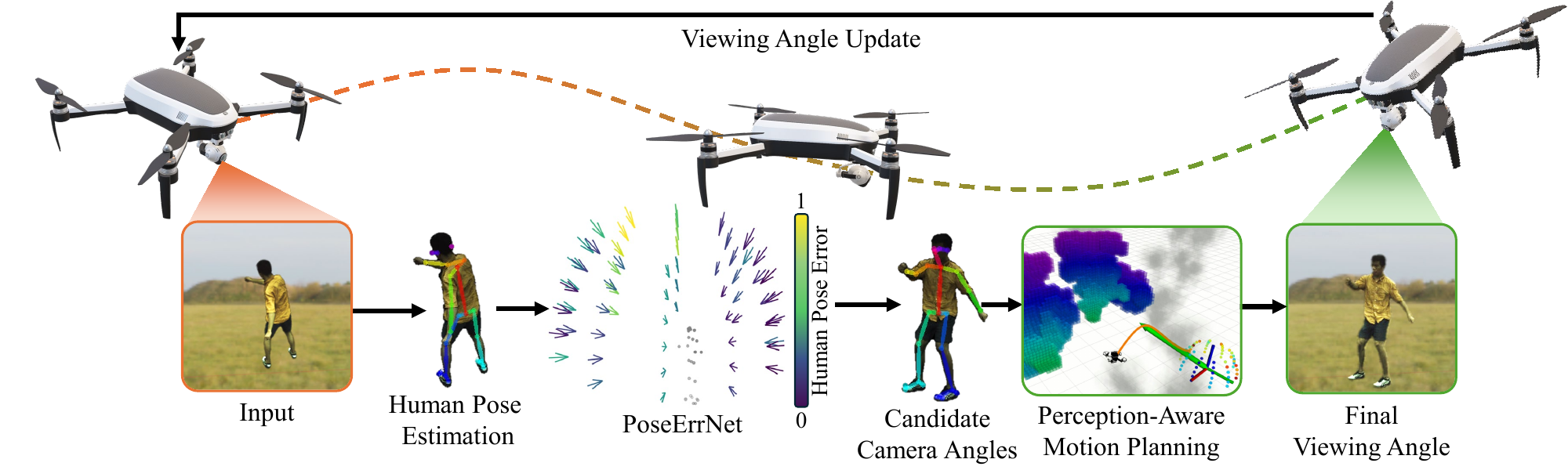}
    \captionsetup{font={small}}
    \caption{
The proposed pipeline for an autonomous drone to perform active human pose estimation follows a specific loop. Initially, the drone captures an input image, which is processed to perform 2D human pose estimation, yielding an imperfect 2D skeleton. Utilizing this skeleton, our designed PoseErrNet predicts the pose estimation error across a hemispherical space of camera viewing angles and suggests candidate camera viewing angles for the best next view. These candidate views are subsequently integrated into a planner focusing on perception-aware planning and navigation goal planning. }
    \label{fig:Banner}
    \end{figure}
}
\maketitle

\addtocounter{figure}{-1} 

\thispagestyle{plain}
\pagestyle{plain}

\begin{abstract}

One of the core activities of an active observer involves moving to secure a "better" view of the scene, where the definition of "better" is task-dependent. This paper focuses on the task of human pose estimation from videos capturing a person's activity. Self-occlusions within the scene can complicate or even prevent accurate human pose estimation. To address this, relocating the camera to a new vantage point is necessary to clarify the view, thereby improving 2D human pose estimation. This paper formalizes the process of achieving an improved viewpoint. Our proposed solution to this challenge comprises three main components: a NeRF-based Drone-View Data Generation Framework, an On-Drone Network for Camera View Error Estimation, and a Combined Planner for devising a feasible motion plan to reposition the camera based on the predicted errors for camera views. The Data Generation Framework utilizes NeRF-based methods to generate a comprehensive dataset of human poses and activities, enhancing the drone's adaptability in various scenarios. The Camera View Error Estimation Network is designed to evaluate the current human pose and identify the most promising next viewing angles for the drone, ensuring a reliable and precise pose estimation from those angles. Finally, the combined planner incorporates these angles while considering the drone's physical and environmental limitations, employing efficient algorithms to navigate safe and effective flight paths. This system represents a significant advancement in active 2D human pose estimation for an autonomous UAV agent, offering substantial potential for applications in aerial cinematography by improving the performance of autonomous human pose estimation and maintaining the operational safety and efficiency of UAVs.

\end{abstract}

\section{Introduction}
\label{sec:intro}

Recent advances in aerial robotic technologies have significantly improved the use and abilities of aerial robots in many industries. \cite{stocker2017review, nex2014uav, baiocchi2013uav}. A key aspect of modern aerial robots is their ability to be equipped with video cameras, transforming them into dynamic platforms for aerial videography \cite{zhou2022swarm}. The mobility and agility of aerial robots make them highly effective for aerial cinematography, allowing for versatile footage capture from optimal angles with minimal equipment. The growing demand in entertainment, industrial, and military sectors has shifted aerial cinematography's focus from static objects to dynamic human subjects, and the technical challenge is how to autonomously adjust a drone's viewing direction to best view human subjects during navigation.

The challenges associated with autonomously adjusting the viewing direction for UAV-based human pose estimation include determining the criteria for modifying the UAV's viewing direction during human inspection, and navigating the UAV in a way that balances perceptual guidance with navigation objectives, such as feasible motion plans and collision avoidance.

\begin{figure*}[t]
    \centering
    \includegraphics[width=0.85\textwidth]{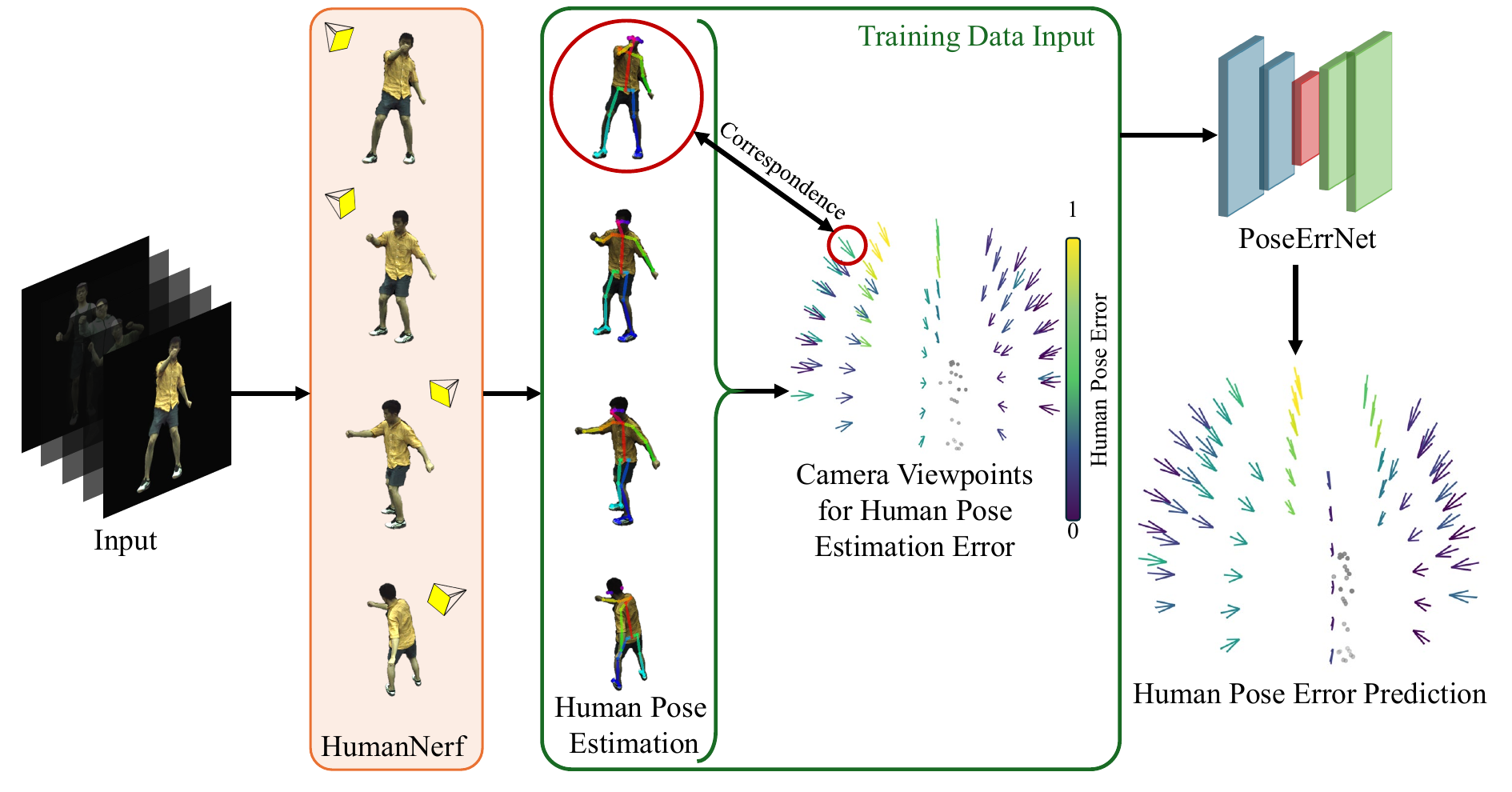}
    \captionsetup{font={small}}
    \caption{ 
Our proposed approach features an integrated system with three key components: 1) Drone-View Data Synthesis, which generates realistic drone perspectives of human subjects from various camera angles and human poses, alongside calculating the associated human pose estimation error for these views to serve as training data pairs. 2) PoseErrNet, a network trained on the generated drone-view data pairs, is capable of predicting a 3D perception guidance field for the selection of candidate viewing angles. 3) A comprehensive planner that integrates traditional navigation cost maps with the 3D perception guidance field derived from PoseErrNet. This integration enables effective motion planning, collision avoidance, and the execution of the next-best viewing angle selection for accurate human pose estimation.   }
    \label{fig:pipeline}
\end{figure*}

To address the challenges posed by dynamic videography using UAVs, we propose a sophisticated, integrated autonomous UAV videography system, as illustrated in Fig. \ref{fig:pipeline}. This system is engineered to intelligently interpret human poses and proactively reposition itself to capture optimal visual content. The architecture of this system can be divided into three primary components: 1) Drone-View Human Subject Data Generation Framework. This framework is designed to capture a wide range of human poses and actions under varying environmental perspectives. By utilizing advanced vision techniques (HumanNerf) \cite{weng_humannerf_2022_cvpr}, this framework will enable the UAV to have a profound understanding of human subjects in different scenarios, enhancing its ability to adapt to real-world videography tasks. 2) Robust and Efficient On-Drone Network for Viewing Angle Estimation. This network is tailored to analyze the current human pose and compute the best subset of the next viewing angles. It aims to process complex visual inputs in real-time, ensuring the UAV can react promptly and accurately to dynamic subjects. The efficiency of this network is crucial, as it directly impacts the UAV's ability to operate under computational and power constraints typically associated with autonomous drones. 3) Combined Planner for Feasible Motion Plan. Our proposed system is a sophisticated planning module that combines the network's viewing angle recommendations with the UAV's dynamic and environmental constraints. The planner employs advanced algorithms to chart a feasible motion plan that not only adheres to the suggested viewing angles but also respects the physical limitations of the UAV and the navigational challenges posed by the environment. Using this planner, the UAV can maneuver in complex environments with agility and precision, ensuring high-quality videography while maintaining safety and operational efficiency.

In Sec. \ref{sec:Drone_data_gen}, we will delve into the specifics of generating human subject data from drone views. Following that, in Sec. \ref{sec:mapping}, we explain our approach using PoseErrNet to transform an imperfect detection of 2D human keypoints, into an error vector for 2D human pose estimation (HPE) across all predefined hemispherical camera viewing angles.  This process creates what we refer to as the 3D perception guidance field. Our goal was to design a lightweight network capable of learning the correlation between the optimal subset of next viewing angles and the current human pose estimation, utilizing a dataset we generated for this purpose. The robust estimation of the 3D perception guidance field is crucial as it provides candidate camera viewing angles for the ensuing motion planning phase.
For the motion planning part, based on \cite{Zhou2020EGOPlannerAE} and \cite{wang2022gpa}, we crafted a perception-aware motion planning framework.  This framework not only incorporates the 3D perception guidance field but also is capable of generating a smooth flight trajectory, avoiding occlusions between the target and the UAV, and ensuring the safety of the flight.

Our contributions can be summarized as:
\begin{itemize}
    \item A drone-view data generation framework for different human poses. 
    \item A robust and efficient network running on the drone for estimating the best subset of the next viewing angles based on the current human pose estimation.
    \item A combined planner that combines the perspective-aware guidance from the network and traditional navigation constraints into a feasible motion plan for improving 2D HPE, a computer vision task.
\end{itemize}

These three interconnected components are seamlessly integrated into a system designed for application in real-world scenarios.

\section{Related Works}
\label{sec:related_work}

\subsubsection{Autonomous Aerial Human Inspection} Existing research on autonomous aerial inspection of human subjects primarily aims at achieving planning autonomy but often lacks objective guidance on subsequent movements. As a result, high-level guidance for the inspection tasks is typically expected to come from human operators, as seen in various studies \cite{zhang2022auto, joubert2015interactive, joubert2016towards, gebhardt2016airways, gebhardt2018optimizing, lan2017xpose, kang2018flycam}. While the autonomous planning for UAVs to follow and robustly track mobile objects along optimized trajectories is well-documented \cite{jeon2019online, elastictracker, jeon2020integrated, wang2021visibility},  these works generally do not address the capability of UAVs to autonomously execute perception-aware objectives, such as human pose estimation, without human operator inputs.

\subsubsection{2D Human Pose Estimation} Advancements in deep learning techniques have significantly enhanced the performance of 2D Human Pose Estimation (HPE), leading to robust and efficient solutions for both single and multiple individuals. \cite{nie2018human, chu2017multi, newell2016stacked, peng2018jointly, toshev2014deeppose, 8765346}. However, the near-perfect performance of 2D Human Pose Estimation (HPE) often relies on ideal input images of humans without any occlusion of body parts. This assumption becomes challenging in the context of autonomous UAV inspections of humans. As the UAV moves, the camera's view of human subjects can easily be obstructed by environmental elements or self-occlusion of human body parts.

\begin{table*}[htb]
    \centering
    \resizebox{0.9\linewidth}{!}{
    \begin{tabular}{@{}cccccccccccc@{}}
    \toprule
     \multirow{2}{*}{\textbf{Method}} &  \multicolumn{2}{c}{\textbf{Cost}}  & \multicolumn{2}{c}{\textbf{Viewing Angle}} & \multicolumn{2}{c}{\textbf{Realistic Capture}}   \\ 
     & Ecnomoic & Engineering & Accuracy & Resolution & Appearance & Human Pose \\ 
        \midrule
        Drone Capture  & Medium   & High &  Low &  Low &  \checkmark &   \checkmark\\
        Camera Array  & High  & High & High & High & \checkmark &   \checkmark \\
        Simulation Software  & Low  & High & High & High &  --  &  -- \\
        \toprule
         Ours & Low & Low  & High & High &  \checkmark &  \checkmark \\
        \toprule
    \end{tabular}
    }
    \captionsetup{font=small}
    \caption{Comparison of different drone-view data acquisition methods. Cost, related to economic and engineering cost of the capture method (\textit{Density}). Viewing Angle, consists of viewing angle accuracy and resolution of the capture method (\textit{Viewing Angle}). Realistic Capture, is related to whether the capture method can capture realistic appearance and human pose (\textit{Realistic Caprure).}}
    \label{tab:datamethods}
\end{table*}

\subsubsection{Neural Radiance Field} NeRF \cite{mildenhall2021nerf} and its extensions \cite{barron2021mip, hedman2021baking, niemeyer2021giraffe, srinivasan2021nerv, tancik2020fourier, zhang2020nerf++, zhang2021nerfactor} enable high-quality and continuous rendering of static 3D scenes. A natural progression is to expand the neural radiance field approach to encompass dynamic scene representation.\cite{gao2021dynamic, li2021neural, park2021hypernerf, pumarola2021d, tretschk2021non, xian2021space}.  In the context of dynamic scene representations, our work is most closely related to the neural representation of dynamic human subjects \cite{weng_humannerf_2022_cvpr, su2021nerf, xu2021h}.

\begin{figure}[t!]
    \centering
    \includegraphics[width=\linewidth]{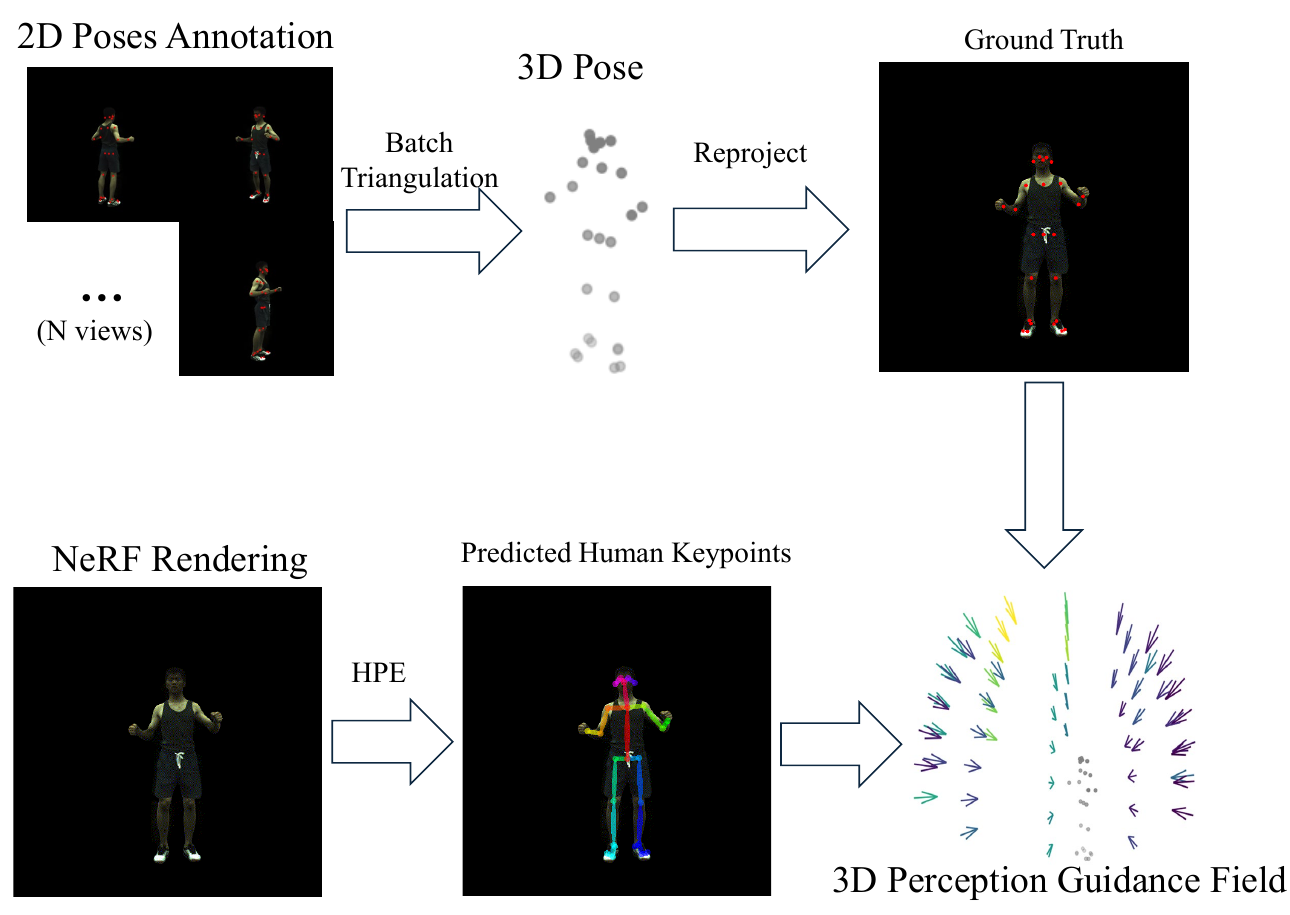}
    \captionsetup{font={small}}
    \caption{
The process for generating NeRF-based drone-view images of human subjects and 3D perception guidance field data involves using 2D annotations to conduct batch triangulation, resulting in a 3D skeleton for a given human pose. We then render the synthesized image for "drone views", reproject the ground truth 3D skeleton onto NeRF poses to obtain ground truth 2D keypoints, and employ an arbitrary HPE network to predict these keypoints for computing the per camera view HPE error. Through this method, we successfully acquire paired data comprising 2D observations and the corresponding 3D perception guidance field. }
    \label{fig:data_gen}
\end{figure}

\section{Drone-View Data Acquisition}
\label{sec:Drone_data_gen}
As illustrated in Fig. \ref{fig:pipeline}, the drone views of a human subject can be represented in a hemispherical space. This type of data can be acquired through one of three methods: 1) Drone Capture, which involves using UAVs to obtain images of human subjects in specific poses from multiple angles. 2) Camera Array, which entails setting up an array of cameras to cover the hemispherical space, with a focus on achieving time synchronization among the cameras. 3) Utilization of simulation software like Blender \cite{Blender} or Unity \cite{Unity} to create projected image views from human models. Each method comes with its own set of challenges and practical considerations. The pros and cons of these methods for generating drone-view data are detailed in Table \ref{tab:datamethods}, highlighting economic and engineering costs. For instance, deploying multiple drones for data capture or creating a camera array incurs significant economic costs due to the hardware required and demands considerable engineering effort for calibration and synchronization. Conversely, simulation software offers a low economic cost option, though achieving a high-quality simulation presents substantial engineering challenges. The table also compares the accuracy of the desired viewing angles, the resolution, or the detail level at which capture angles are set, and the realism of capturing human subjects and poses. Drone capture and camera arrays provide realism in both appearance and pose since they employ real-world methods. In contrast, achieving realistic captures of both appearance and human poses proves difficult with simulation software.

Our research leverages the innovative free-viewpoint rendering method, HumanNeRF \cite{weng_humannerf_2022_cvpr}, which is designed for rendering images with complex human poses, perfectly meeting our data acquisition needs. This technique allows for the free-view synthesis of a human image in a specific pose. We configure the camera poses and viewing angles to match the desired hemispherical drone camera pose for the captured image and then render drone-view images for various human poses.

After rendering the drone-view images, we follow the approach depicted in Fig. \ref{fig:data_gen} to compute the human pose estimation error for each camera view. This process enables us to gather the desired training data pairs, linking each 2D human skeleton estimation to a 3D perception guidance field.

\begin{figure*}[t!]
    \centering
    \includegraphics[width=\linewidth]{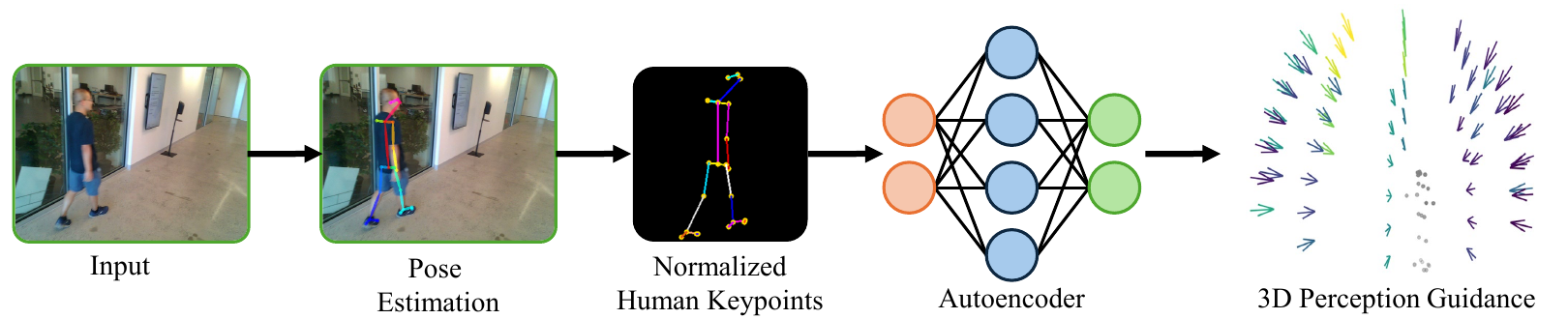}
    \captionsetup{font={small}}
    \caption{PoseErrNet: It consists of two major parts: 1) Input abstraction and normalization to deal with the sim-to-real gap and with scale, translation, and rotation invariance for the drone applications, and 2) The auto-encoder network to map from normalized 2D observations to 3D perception guidance fields.}
    \label{fig:method}
\end{figure*}

\section{Mapping 2D Observations to 3D perception guidance fields}
\label{sec:mapping}
After acquiring our training data pairs for drone-view human images and 3D perception guidance field.  A simplified auto-encoder network is employed for visual guidance, as depicted in Fig. \ref{fig:method}. This network architecture is characterized by a minimal number of weights, enhancing its efficiency. For dealing with the sim-to-real gap, we proposed a process to normalize the input drone-view data, this normalization process includes first converting the input image into the HPE resulting keypoints and then normalizing the detected keypoints to account for translation, rotation, and scale variance in the input keypoint due to different human, drone-to-human distance or instability during the flight of the drone.

The proposed normalization process for input HPE keypoints is straightforward yet effective. We begin by identifying the human spine, typically the line between the neck and the midpoint of the hips. Once the spine is determined, we translate all keypoints to align the spine's midpoint to a consistent coordinate, addressing the translation variance of the input keypoints. Furthermore, we rotate all keypoints to orient the spine's direction upwards and to the right, countering the rotation variance of the input keypoints. Finally, we scale all keypoints to ensure the spine length remains constant. An example of normalized human keypoints is illustrated in Fig. \ref{fig:method}.

Our normalization process for the HPE keypoints enhances robustness against variances in translation, rotation, and scale in the inputs.  By using the normalized coordinates of the keypoints, represented as a vector, as input to our network, we simplify the design of the network architecture. The proposed PoseErrNet is an autoencoder with a minimal number of layers, benefiting greatly from the simplicity and normalization of its inputs.

\section{Perception-aware motion planning}

\begin{figure}[t!]
    \centering
    \includegraphics[width=\linewidth]{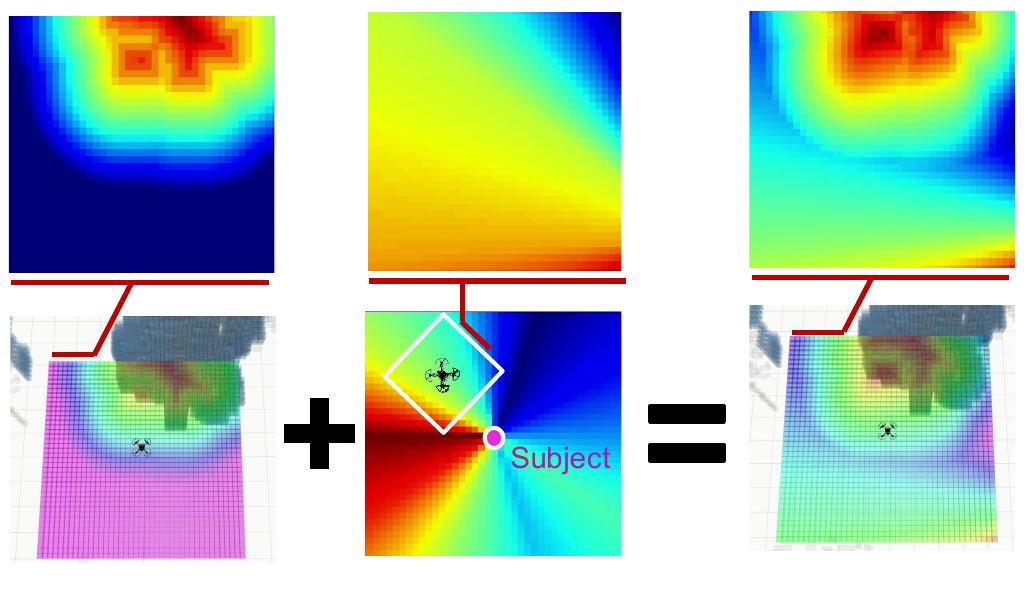}
    \captionsetup{font={small}}
    \caption{Visualization for the calculation of P-ESDF.}
    \label{fig:pesdf}
\end{figure}

 We introduce our proposed perception-aware motion planning framework to bridge the gap between perception and motion planning. In the proposed framework, the perception loss is added to the motion planning cost function as one of the costs. To achieve that, we built a differentiable distance field called Pose-enhanced Euclidean Distance Field (P-ESDF), noted as $\mathcal{P}$, each of its element is then represented as $\{ p_i \in \mathcal{P} | i \in \mathbb{Z}^+\}$. The construction of the field is described below.
 
 The output of the Pose Estimation Error Predictor is a 2-D map $\mathcal{E}: \mathbb{R}^{m \times n} \rightarrow \mathbb{R}$.
 The first step is to transform $\mathcal{E}$ to the subject frame in $\mathbb{R}^3$ with the subject in the center, as shown in the bottom sub-figure of Fig. \ref{fig:pesdf}(b). Then, we project the transformed map $\mathcal{E}^{sub}$ from the subject frame to the drone frame to get the $\mathcal{E}^{drone}$, as shown in the upper sub-figure of Fig. \ref{fig:pesdf}(b).
To simutanrously perform obstacle avoidance and viewpoint targeting, the $\mathcal{E}^{drone}$ need to be merged with the standard ESDF $\mathbf{E}$ to get the final $\mathcal{P}$, as illustrated in Fig. \ref{fig:pesdf}. The merge process can be expressed as:

\begin{equation}
    \mathcal{P} = \sum_{i} \lambda \mathcal{E}^{drone}_{i} + (1-\lambda) \mathbf{E}_{i}.
\end{equation}

Then, to guide the drone with the proposed P-ESDF, we design pose penalty $J_{pose}$ as the function of $p$. Assume the path is constructed by a series of waypoints $\{ \mathbf{p_k} \subset \mathcal{R}^3 | k \in \mathbb{Z}^+\}$, and define $\Xi(\mathbf{p_k})$ as the value of $\mathcal{P}$ at the position of $\mathbf{p_k}$.
The $J_{pose}$ can be expressed as

\begin{equation}
    J_{pose} = \lambda_p \sum_{i=0}^{M} c(p_i) \mathbf{p_k}'
\end{equation}
where $\mathbf{p}_k' = \frac{\partial \Xi(\mathbf{p}_k)}{\partial \mathbf{p}_k}$ can be efficiently acquired from P-ESDF, and $c(\mathbf{p}_k)$ can be expressed as:
\begin{equation}
    c(\mathbf{p}_k) = 
    \begin{cases}
    \displaystyle\frac{1}{2\rho}(\Xi(\mathbf{p}_k)-\rho)^2, \Xi(p_k) \leq \rho \\
            0\hfill, \Xi(\mathbf{p}_k) > \rho
    \end{cases}
\end{equation}

The result is shown in Fig. \ref{fig:pesdf}.

\section{Experiments}

To demonstrate the effectiveness of our system from different dimensions, we conduct $3$ tasks in simulated environments with varied scales and complexity, as shown in Fig. \ref{fig:env}. 
The first task is to estimate a static challenging pose, as shown in Fig. \ref{fig:env} (a).
The later two tasks are to estimate the human pose online during walking in the forests, as shown in Fig. \ref{fig:env} (b-c). The drone needs to simutaneously track the person, choose the best view-point, avoid occlusion and ensure the safety. 

\subsection{Implementation Details}
\label{sec:ImpleDetail}

The implementation of PoseErrNet is an autoencoder with 4 layers in the encoder layers and 4 layers in the decoder layers. 
The TRT-Pose is utilized for 2-D human pose estimation. It runs on NVIDIA RTX 3070Ti GPU with 15 HZ update rate.
The simulated vehicle is equipped with an Intel D435 depth camera, which is used both as the range sensor for navigation planning and camera sensor for RGB images. 
The onboard autonomy system of the UAV integrates several key navigation modules from the development environment of \cite{autofilmer}. These include kinodynamic path search, mapping module and GUI. These components serve as fundamental navigation modules. The proposed perception-aware planner is on the top of the navigation system.
The framework runs on a laptop with i7-12700H CPU. We configure the navigation system to update at 15Hz and perform trajectory optimization at each sensor update. The spatial resolution is set as $0.2m$. The P-ESDF is set a $10 \times 10 m$ area with the vehicle in the center.

\subsection{Evaluation of 3D Perception Guidance Field Generation}
\label{sec:EvaPerception}

\begin{table*}[t!]
    \centering
    \resizebox{0.6\linewidth}{!}{
    \begin{tabular}{@{}cccccccccccc@{}}
    \toprule
     \multirow{1}{*}{\textbf{Perturbation}} &  \multicolumn{1}{c}{\textbf{Translation (\%)}}  & \multicolumn{1}{c}{\textbf{Rotation  (\%)}} & \multicolumn{1}{c}{\textbf{Scale  (\%)}}  & \multicolumn{1}{c}{\textbf{All  (\%)}}  \\ 
     
        \midrule
        T1  & 5   & 8 &  12 & 17\\
        T2  & 11  & 12 & 16 & 25  \\
        T3  & 19  & 18 & 21 & 34  \\
        \toprule
    \end{tabular}
    }
    \captionsetup{font=small}
    \caption{Evaluation of Robustness of Pose Normalization for 3 perturbation levels T1, T2, and T3. The translation-only results (\textit{Translation}),  the rotation-only results (\textit{Roataion}),  the scale-only results (\textit{Scale}) and combing all translation, rotation, scale results (\textit{All}). Here we show after input keypoint normalization how many percentages of PoseErrNet output change due to input keypoint perturbation.  }
    \label{tab:robustness}
\end{table*}

\begin{figure}[t!]
    \centering
    \includegraphics[width=0.8\linewidth]{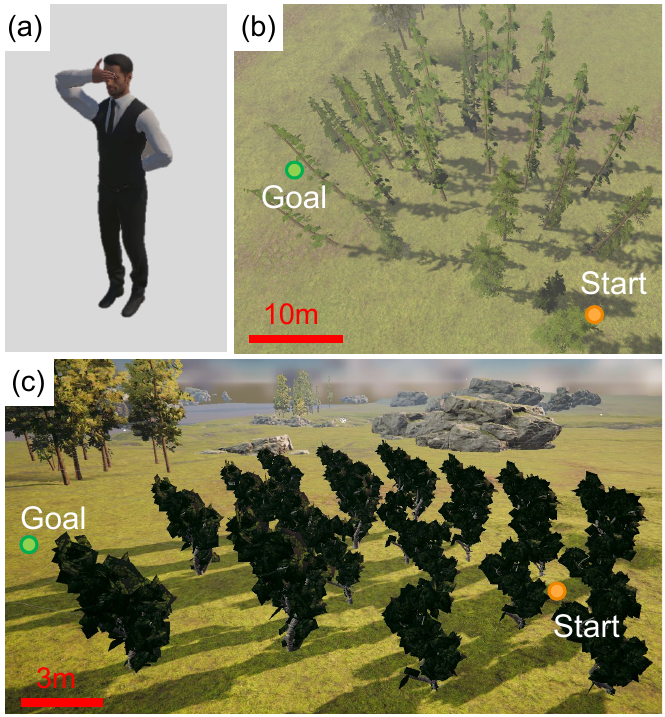}
    \captionsetup{font={small}}
    \caption{The testing environment for the proposed motion planning framework.}
    \label{fig:env}
\end{figure}

\begin{figure}[t!]
    \centering
    \includegraphics[width=1\linewidth]{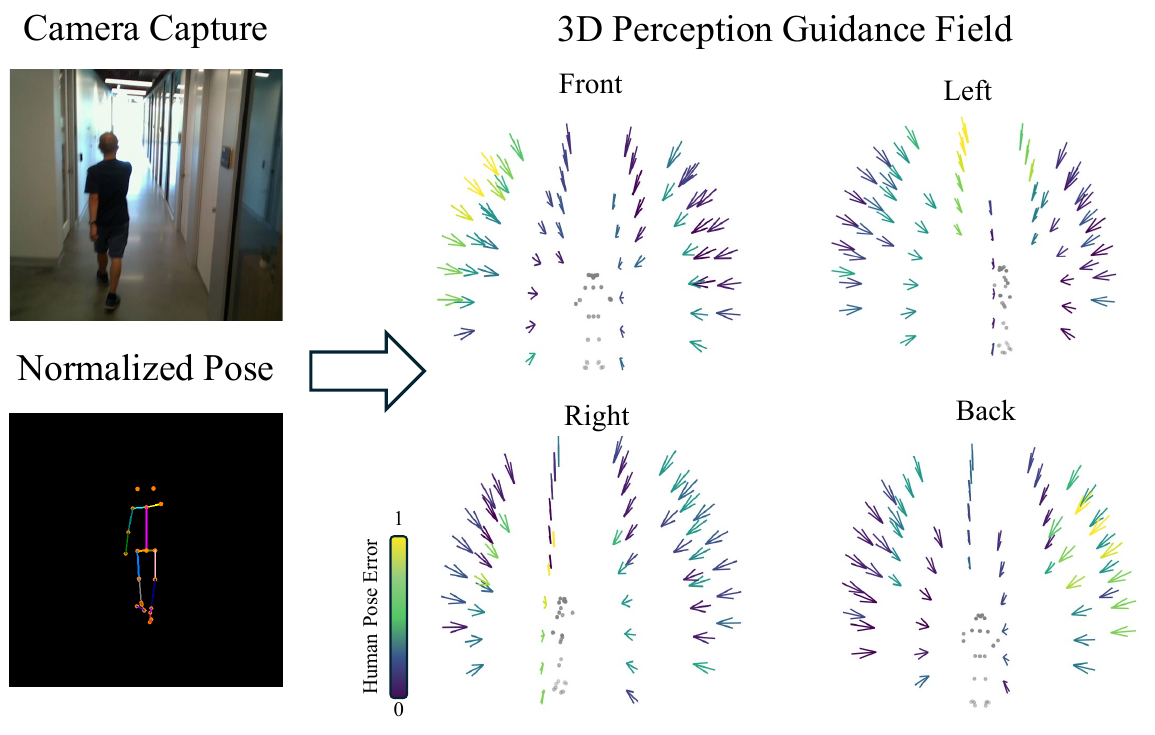}
    \captionsetup{font={small}}
    \caption{The result for 3D perception guidance field prediction on the example real-world data.  Each arrow within the 3D perception guidance field represents a camera viewing angle. Colors are utilized to denote the error of HPE at a camera viewing angle. }
    \label{fig:ResPercep}
\end{figure}

To demonstrate the performance of the proposed 3D perception guidance field generation, we showcase an example in challenging real-world scenarios as shown in Fig. \ref{fig:ResPercep}.

In Fig. \ref{fig:ResPercep}, the algorithm begins by normalizing the detected imperfect 2D human keypoints, as outlined in the methodology section (Sec.~\ref{sec:mapping}).  The robustness of our PoseErrNet's output against variations in scale, translation, and rotation of the input keypoint detection is enhanced by this normalization process. We captured a real-world video with a hand-held camera following a human subject and selected 146 representative frames. To these frames, we applied three levels of perturbation: T1 involves uniform random translation ranging from 0 to 5 pixels, uniform random rotation from 0 to 5 degrees, and uniform random scaling from 1.0 to 1.05; T2 includes uniform random translation from 0 to 10 pixels, uniform random rotation from 0 to 10 degrees, and uniform random scaling from 1.0 to 1.10; T3 consists of uniform random translation from 0 to 20 pixels, uniform random rotation from 0 to 20 degrees, and uniform random scaling from 1.0 to 1.15. As illustrated in Table \ref{tab:robustness}, we quantized the camera viewing angle error into 21 bins within its range and reported the percentage of bin changes from the results without perturbation for all camera viewing angle error predictions from PoseErrNet. Due to the input normalization, our PoseErrNet's output demonstrates robustness under various levels of input keypoint perturbations.

This normalized keypoints detection data is then input into PoseErrNet to predict the 3D perception guidance field. As an example in Fig. \ref{fig:ResPercep} from our real-world collected video frames,  where a person moves forward while raising their right arm, obstructing their face. Viewpoints directly in front of the individual are assigned lower costs due to their superior visibility. In contrast, viewpoints from behind are usually associated with higher costs, as they are more prone to occluding important features like the face and arms. The error increases on the right-hand side due to occlusion caused by the raised arm. The left to the front side also with high error because the person's left leg moving forward creates self-occlusion of the right leg. Meanwhile, the area from the left to the back side exhibits the lowest error, indicating the best candidate camera viewing angles for this scenario.

\subsection{Perception-aware Motion Planning Experiment}
\label{sec:EvaPlanning}

To evaluate the efficacy of the On-drone Perception-aware Motion Planning method, we conducted an experimental implementation in complex, cluttered environments characterized by numerous obstacles, as depicted in Fig. \ref{fig:env}. 

\begin{figure}[ht!]
    \centering
    \includegraphics[width=\linewidth]{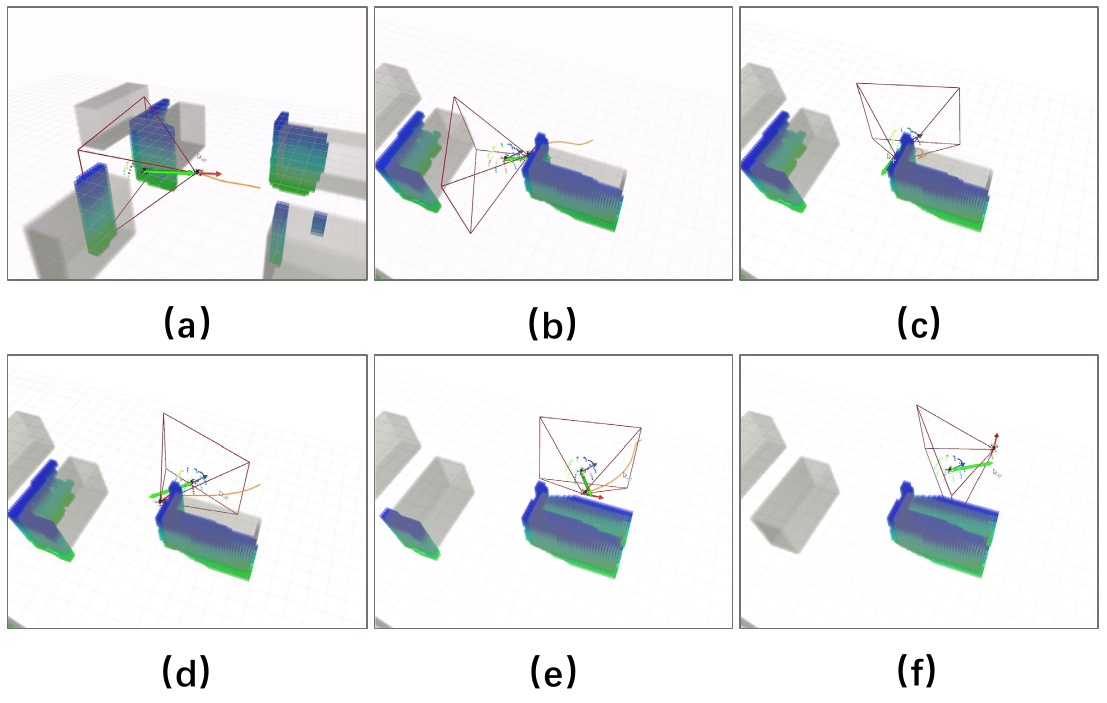}
    \captionsetup{font={small}}
    \caption{Perception-aware motion planning experiment. \textbf{(a): }The UAV follows the best viewpoint (green arrow). \textbf{(b): }The UAV can still follow the best viewpoint and while avoiding the obstacle. \textbf{(c-d): }To avoid the occlusion between the target and the UAV, and also to avoid the obstacles for safety, the proposed planning framework automatically switches to the second best viewpoint. \textbf{(e-f): }After avoiding the obstacle, and if there is no occlusion between the target and the UAV, the UAV plans a smooth trajectory to return to the best viewpoint.}
    \label{fig:ObsAvoid}
\end{figure}

As shown in Fig. \ref{fig:sim} and Fig. \ref{fig:ObsAvoid}, in obstacle-free environments, the UAV consistently adheres to the optimal viewpoint. However, upon encountering obstacles, the UAV demonstrates the capability to maintain this optimal viewpoint while simultaneously navigating around the obstructions. In scenarios where obstacles are proximate or densely situated, the planning framework proactively shifts to the second-best viewpoint. This adjustment is crucial for mitigating occlusion issues between the target and the UAV, and for maintaining safety by avoiding obstacles. Upon successful navigation past  obstacles, and making sure  that occlusions do  not obstruct the view of the target, the UAV seamlessly plans and executes a trajectory to return to the primary, most advantageous viewpoint. 

\subsection{System-level Comparison}
\label{sec:SysExp}

\begin{figure*}[t!]
    \centering
    \includegraphics[width=\textwidth]{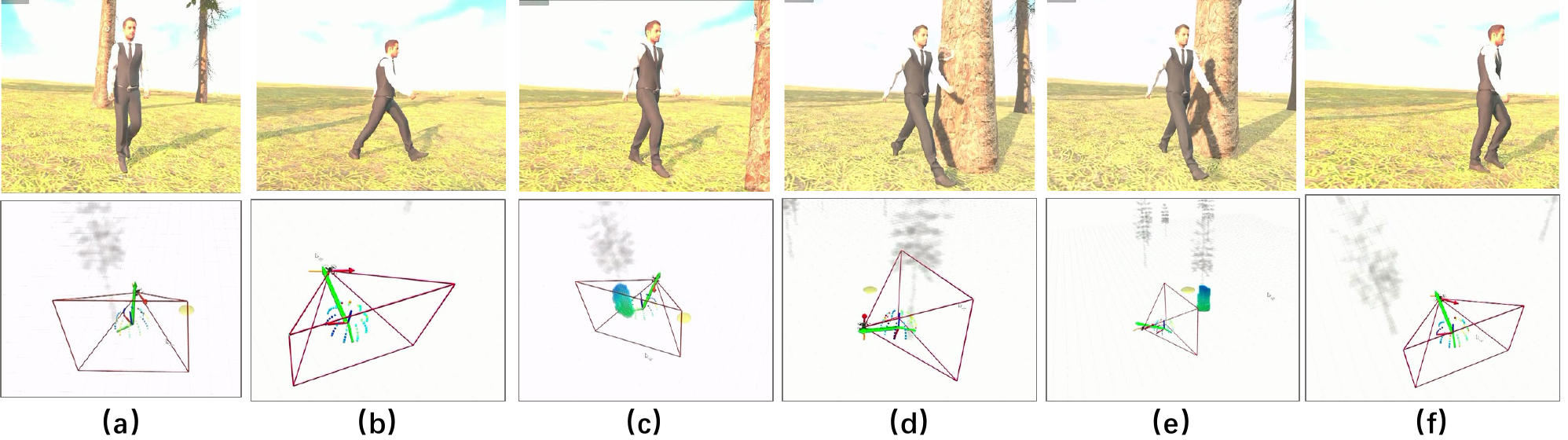}
    \captionsetup{font={small}}
    \caption{Demonstration of the perception-aware motion planning. The upper row in each figure is an image captured using the onboard camera, and the bottom row illustrates  the third-person view of the experiment. The 3D perception guidance field is represented as a hemispherical field around the drone with different colors indicating the quality of the view point (from green to red indicating good to poor). The green arrow means the best viewpoint evaluated altogether from the pose estimation accuracy, occlusion, and flight safety. \textbf{(a - b): }The best viewpoint (green arrow) changes with  the human pose, and the UAV follows the best view point in real-time.  \textbf{(c - f): }When  both target and obstacles are observed in the environment, the UAV automatically chooses the second best viewpoint to avoid occlusions and ensure safety.}
    \label{fig:sim}
\end{figure*}

\begin{table*}[]
\centering
\resizebox{0.8\linewidth}{!}{%
\begin{tabular}{c|cccccccc}
\hline
\multirow{2}{*}{\textbf{Method}} & \multicolumn{2}{c}{\textbf{Challenging Pose}} & \multicolumn{2}{c}{\textbf{Large-Scale}} & \multicolumn{2}{c}{\textbf{Dense}} & \multicolumn{2}{c}{\textbf{All}} \\
                                 & PCK                  & MSE                    & PCK                & MSE                 & PCK             & MSE              & PCK            & MSE             \\ \hline
Auto-Filmer - Front              & 0.87                 & 31.46                  & \textbf{0.91}      & \textbf{24.10}      & 0.60            & 42.41            & 0.79           & 32.66           \\
Auto-Filmer - Side               & 0.93                 & 30.73                  & 0.78               & 25.59               & 0.80            & 37.89            & 0.84           & 31.40           \\
Auto-Filmer - Back               & 0.80                 & 29.20                  & 0.67               & 34.31               & 0.67            & 37.51            & 0.71           & 33.67           \\ \hline
Ours                             & \textbf{1.0}         & \textbf{22.60}         & 0.86               & 24.47               & \textbf{0.92}   & \textbf{23.61}   & \textbf{0.92}  & \textbf{23.56}  \\ \hline
\end{tabular}%
}
\caption{PCK and MSE Evaluation for System-level Experiments}
\label{tab:sys}
\end{table*}
This experiment was designed to validate the robustness and overall performance of our integrated system. The perception aspects were simulated within Unity, while the control and motion planning components were simulated in ROS \cite{Quigley2009ROSAO} using Rviz \cite{094f52e6f6804ec09c1b4e0b4f96a445} for visual representation.
Two standard metrics are introduced to evaluate the performance: Percentage of Correct Key-points (PCK) and Mean Squared Error.

As indicated in Fig. \ref{fig:sim} and Table. \ref{tab:sys}, the optimal viewpoint (represented by a green arrow) dynamically adjusts in response to changes in the human subject's pose. The UAV is programmed to track and align with this best viewpoint in real-time, showcasing its responsiveness to the target's movements. In scenarios where both the target and obstacles are present within the UAV's operational environment, the system intelligently opts for the second-best viewpoint. This strategic choice is critical for avoiding visual occlusion between the UAV and the target, and for ensuring safety by steering clear of obstacles. This approach effectively demonstrates the system's capability to adapt to varying environmental conditions while maintaining high-quality perception and safe navigation.

\section{Conclusion}

The innovative approach detailed in this paper signifies a considerable leap forward in the domain of active 2D human pose estimation through the use of autonomous Unmanned Aerial Vehicles (UAVs). By weaving together a NeRF-based Drone-View Data Generation Framework, an On-Drone Network for Camera View Error Estimation, and a Combined Planner for strategic camera repositioning, our methodology effectively tackles the issue of self-occlusions in videos capturing human activities. This integrated system not only enhances the accuracy and reliability of human pose estimation from optimized camera viewing angles but also guarantees the adaptability and operational safety of drones across varied environments. The proposed method highlights the critical role of dynamic viewpoint optimization in elevating the quality of pose estimation, thereby paving new pathways for applications in sectors like aerial cinematography and surveillance. Experimental results from both simulation and real-world experiments prove the efficacy of each component within our system and, more importantly, demonstrate the enhanced task-level performance of the integrated system.

\bibliographystyle{unsrt}
\bibliography{main}

\end{document}